\documentclass{scrartcl}
\usepackage{hyperref}
\usepackage{booktabs}
\usepackage{tikz}
\usepackage{python}
\usepackage{libertine}
\usepackage{DejaVuSansMono}
\usepackage{natbib}
\hypersetup{%
    bookmarks=true,
    colorlinks=true,
    pdftitle={Mahotas: Open source software for scriptable computer vision},
    pdfauthor={Luis Pedro Coelho},
    citecolor=pondgreen,
    urlcolor=toastedchilipowder,
    linkcolor=toastedchilipowder,
}

\newcommand*{\cpp}{{C\nolinebreak[4]\hspace{-.05em}\raisebox{.4ex}{\tiny\textbf{++}}}}
\let\code\texttt
\title{Mahotas: Open source software for scriptable computer vision}
\author{Luis Pedro Coelho\\
Lane Center for Computational Biology, Carnegie Mellon University\\
Instituto de Medicina Molecular}
\date{January 2013}

\begin{document}
\maketitle

\section*{Abstract}
Mahotas is a computer vision library for Python. It contains traditional image
processing functionality such as filtering and morphological operations as well
as more modern computer vision functions for feature computation, including
interest point detection and local descriptors.

The interface is in Python, a dynamic programming language, which is very
appropriate for fast development, but the algorithms are implemented in \cpp{}
and are tuned for speed. The library is designed to fit in with the scientific
software ecosystem in this language and can leverage the existing
infrastructure developed in that language.

Mahotas is released under a liberal open source license (MIT License) and is
available from \url{http://github.com/luispedro/mahotas} and from the Python Package
Index (\url{http://pypi.python.org/pypi/mahotas}).

\textbf{Keywords:} computer vision, image processing.

\section{Introduction}

Mahotas is a computer vision library for the Python Programming Language
(versions 2.5 and up, including version 3 and up). It operates on numpy
arrays~\citep{numpystructure}. Therefore, it uses all the infrastructure built
by that project for storing information and performing basic manipulations and
computations. In particular, unlike libraries written in the C~Language or in
Java~\citep{Marcel:2010:TMP:1873951.1874254}, Mahotas does not need to define a
new image data structure, but uses the numpy array structure. Many basic
manipulation functionality that would otherwise be part of a computer vision
library are handled by numpy. For example, computing averages and other simple
statistics, handling multi-channel images, converting between types (integer
and floating point images are supported by mahotas) can all be performed with
numpy builtin functionality. For the user, this has the additional advantage
that they do not need to learn yet another set of functions.

It contains over 100~functions with functionality ranging from traditional
image filtering and morphological operations to more modern wavelet
decompositions and local feature computations. Additionally, by integrating
into the Python numeric ecosystem, users can use other packages in a seamless
way. In particular, mahotas does not implement any machine learning
functionality, but rather advises the user to use another, specialised package,
such as scikits-learn or milk.

Python is a natural ``glue'' language: it is easy to use state-of-the-art
libraries written in multiple languages~\citep{10.1109/MCSE.2007.58}. Mahotas
itself is a mix of high-level Python and low-level \cpp{}. This achieves a good
balance between speed and ease of implementation.

Version 1.0 of mahotas has been released recently and this is now a mature,
well-tested package (the first versions were made available over 4~years ago,
although the package was not named mahotas then). \footnote{Note for reviewers:
the version currently available (0.9.6) implements all functionality described
in this manuscript. I will release it as version 1.0 when this manuscript is
accepted to coincide with publication. Naturally, any bugs that are found and
reported in the meanwhile will be addressed.} Mahotas runs and is used on
different versions of Unix (including Linux, SunOS, and FreeBSD), Mac OS X, and
Windows.

\section{Implementation and Architecture}

\subsection{Interface}

The interface is a procedural interface, with no global state. All functions
work independently of each other (there is code sharing at the implementation
level, but this is hidden from the user).

The main functionality is grouped into the following categories:

\begin{description}
\item[\textsc{Surf}] Speeded-up Robust Features~\citep{eth_biwi_00517}. This
includes both keypoint detection and descriptor computation.
\item[Features] Global feature descriptors. In particular, Haralick texture
features, Zernike moments, local binary patterns, and threshold adjacency
statistics (both the original~\citep{Hamilton2007} and the parameter-free
versions~\citep{Coelho2010}).
\item[Wavelet] Haar and Daubechies wavelets. Forward and inverse transforms are
supported.
\item[Morphological functions] Erosion and dilation, as well as some more
complex operations built on these. There are both binary and grayscale
implementations of these operators.
\item[Watershed] seeded watershed and distance map
transforms~\citep{felzenszwalb}.
\item[Filtering] Gaussian filtering, edge finding, and general convolutions.
\item[Polygon operations] convex hull, polygon drawing.
\end{description}

Numpy arrays contain data of a specific type, such \code{unsigned 8 bit
integer} or floating point numbers. While natural colour images are typically
8~bits, scientific data is often larger and processing can result in floating
point images. Mahotas works on all datatypes. This is performed without any
extra memory copies. Mahotas is heavily optimised for both speed and memory
usage (it can be used with very large arrays).

There are a few interface conventions which apply to many functions. When
meaningful, a structuring element is used to define neighbourhoods or adjacency
relationships (morphological functions, in particular, use this convention).
Generally, the default is to use a $3 \times 3$ cross as the default if no
structuring filter is given.

When a new image is to be returned, functions take an argument named \code{out}
where the output will be stored. This argument is often much more restricted in
type. In particular, it must be a contiguous array.\footnote{Numpy supports
non-contiguous arrays, which are most often slices into other, larger,
contiguous arrays (e.g., given a $128 \times 128$ contiguous array, one can
build a $64 \times 128$ non-contiguous array by taking every other row).} Since
this is a performance feature (its purpose is to avoid extra memory
allocation), it is natural that the interface is less flexible (accessing a
contiguous array is much more efficient than a non-contiguous one).

\subsection{Example of Use}

Code for this and other examples is present in the mahotas source distribution
under the \texttt{demos/} directory. In this example, we load an image, find
SURF interest points, and compute descriptors.

We start by importing the necessary packages, including numpy and mahotas. We
also use \textit{milk}, to demonstrate how the mahotas output can integrate
with a machine learning package.

\begin{python}
import numpy as np
import mahotas
from mahotas.features import surf
import milk
\end{python}

The first step is to load the image and convert to 8~bit numbers. In this
case, the conversion is done using standard numpy methods, namely
\code{astype}:

\begin{python}
f = mahotas.imread('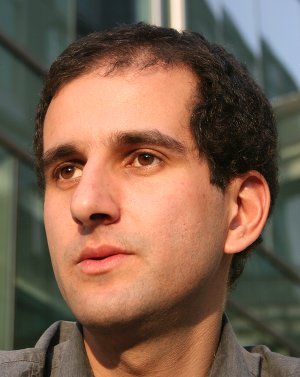', as_grey=True)
f = f.astype(np.uint8)
\end{python}

We can now compute SURF interest points and descriptors.
\begin{python}
spoints = surf.surf(f, 4, 6, 2)
\end{python}

The \code{surf.surf} function returns both the descriptors and their meta data.
We use numpy operations to retain only the descriptors (the meta data is in the
first five positions):

\begin{python}
descrs = spoints[:,6:]
\end{python}

Using milk, we cluster the descriptors into five groups:

\begin{python}
values, _ = milk.kmeans(descrs, 5)
\end{python}

Finally, we can show the points in different colours.
\begin{python}
colors = np.array(
    [ 255,  25,   1],
    [203,  77,  37],
    [151, 129,  56],
    [ 99, 181,  52],
    [ 47, 233,   5]])
f2 = surf.show_surf(f, spoints[:64], values, colors)
\end{python}

\begin{figure}
\begin{center}
\includegraphics[width=.8\textwidth]{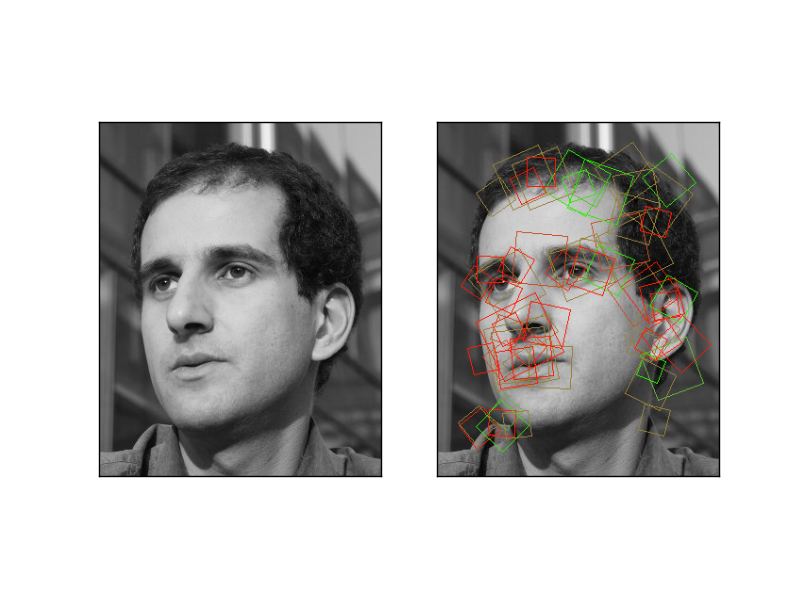}
\end{center}
\caption{Example of Usage. On the left, the original image is shown, while on
the right SURF detections are represented as rectangles of different colours.}
\label{fig:surf}
\end{figure}

The \texttt{show\_surf} only builds the image as a multi-channel (one for each
colour) image. Using matplotlib~\citep{10.1109/MCSE.2007.55}, we finally
display the image as Figure~\ref{fig:surf}.

\begin{python}
from matplotlib import pyplot as plt
plt.subplot(1,2,1)
plt.imshow(f)
plt.subplot(1,2,2)
plt.imshow(f2)
\end{python}

The easy interaction with matplotlib is another way in which we benefit from
the numpy-based ecosystem as mahotas does not need to support interacting with
a graphical system to display images.

\subsection{Implementation}

Mahotas is mostly written in \cpp, but this is completely hidden from the user
as there are hand-written Python wrappers for all functions. Automatically
generated wrappers inevitably lead to worse error messages and are less
flexible.

The main reason that mahotas is implemented in \cpp{} (and not in C, which is
the language of the Python interpreter) is to use templates. Almost \cpp{}
functionality is split across 2~functions:

\begin{enumerate}
\item A \code{py\_function} which uses the Python C~API to get arguments and
check them.
\item A template \code{function<dtype>} which works for the type \code{dtype}
performing the actual operation.
\end{enumerate}

So, for example, this is how \emph{erode} is implemented.\footnote{This is the
generic version of the code. Erode, like a few other functions, has two
versions, a fast version, limited to two-dimensional, contiguous, images and a
generic one presented here. The selection between the two implementations is
done automatically for the user.} \code{py\_erode} consists mostly of
boiler-plate code:

\begin{cplusplus}
PyObject* py_erode(PyObject* self, PyObject* args) {
    PyArrayObject* array;
    PyArrayObject* Bc;
    PyArrayObject* output;
    if (!PyArg_ParseTuple(args, "OOO", &array, &Bc, &output) ||
        !numpy::are_arrays(array, Bc, output) ||
        !numpy::same_shape(array, output) ||
        !numpy::equiv_typenums(array, Bc, output) ||
        PyArray_NDIM(array) != PyArray_NDIM(Bc)
    ) {
        PyErr_SetString(PyExc_RuntimeError, TypeErrorMsg);
        return NULL;
    }
    holdref r_o(output);

#define HANDLE(type) \
    erode<type>(numpy::aligned_array<type>(output), \
                numpy::aligned_array<type>(array), \
                numpy::aligned_array<type>(Bc));
    SAFE_SWITCH_ON_INTEGER_TYPES_OF(array);
#undef HANDLE
    ...
\end{cplusplus}

This functions retrieves the arguments, performs some sanity checks, performs a
bit of initialisation, and finally, switches in the input type with the help of
the \code{SAFE\_\-SWITCH\_\-ON\_\-INTEGER\_\-TYPES} macro, which call the right
specialisation of the template that does the actual work. In this example
\code{erode} implements erosion:

\begin{cplusplus}
template<typename T>
void erode(numpy::aligned_array<T> res,
            numpy::aligned_array<T> array,
            numpy::aligned_array<T> Bc) {
    gil_release nogil;
    const int N = res.size();
    typename numpy::aligned_array<T>::iterator iter = array.begin();
    filter_iterator<T> filter(array.raw_array(), Bc.raw_array(),
                    ExtendNearest, is_bool(T()));
    const int N2 = filter.size();
    T* rpos = res.data();

    for (int i = 0;
            i != N;
                ++i, ++rpos, filter.iterate_both(iter)) {
        T value = std::numeric_limits<T>::max();
        for (int j = 0; j != N2; ++j) {
            T arr_val = T();
            filter.retrieve(iter, j, arr_val);
            value = std::min<T>(value, erode_sub(arr_val, filter[j]));
        }
        *rpos = value;
    }
}
\end{cplusplus}

The template machinery makes the functions that use it very simple and easy to
read. The only downside is that there is some expansion of code size when the
compiler instantiates the function for the several integer and floating point
types. Given the small size of these functions, the total size of the compiled
library is reasonable (circa 6MiB on an Intel-based 64~bit system for the whole
library).

In the snippet above, you can see some other \cpp{} machinery:

\begin{description}
\item[\code{gil\_release}] This is a ``resource-acquisition is object
initialisation'' (\textsc{raii})\footnote{\textsc{Raii} is a design pattern in
\cpp{}, or other languages with scope linked deterministic object destruction,
such as D, where a resource is represented by an object, whose constructor
acquires it and whose destructor releases it. This guarantees that the object
is correctly released even if the scope is left through an exception
\citep{Stroustrup1994}.} object that releases the Python global interpreter lock
(\textsc{gil})\footnote{In the CPython interpreter, the most commonly used
implementation of Python, there is a global lock for many Python related
functionality, which limits parallelism.} in its constructor and gets it back
in its destructor. Normally, the template function will release the
\textsc{gil} after the Python-specific code is done. This allows several
mahotas functions to run concurrently.
\item[\code{array}] This is a thin wrapper around \code{PyArrayObject}, the raw
numpy data type, which has iterators which resemble the \cpp{} standard
library. It also handles type-casting internally, making the code type-safer.
This is also a \textsc{raii} object in terms of managing Python reference
counts. In mahotas debug builds, this object additionally adds several checks
to all the memory acesses.
\item[\code{filter\_iterator}] This was adapted from code in the
\code{scipy.ndimage} packages and it is useful to iterate over an image and use
a centered filter around each pixel (it keeps track of all of the boundary
conditions).
\end{description}

The inner loop is as direct an implementation of erosion as one would wish for:
for each pixel in the image, look at its neighbours, subtract the filter value,
and compute the minimum of this operation.

\subsection{Efficiency}

\begin{table}
\centering
\begin{tabular}{lrrrrrr}
\toprule
Operation            &  mahotas &  pymorph & scikits-image & OpenCV \\
\midrule
erode                &      1.6 &     15.1 &      7.4 &      0.4  \\
dilate               &      1.5 &      9.1 &      7.3 &      0.4  \\
open                 &      3.2 &     24.3 &     14.8 &       NA  \\
median filter (2)    &    226.9 &       NA &   2034.0 &       NA  \\
median filter (10)   &   2810.9 &       NA &   1877.1 &       NA  \\
center mass          &      5.0 &       NA &   3611.2 &       NA  \\
sobel                &     34.1 &       NA &     62.5 &      6.2  \\
cwatershed           &    174.8 &  58440.3 &    287.3 &     44.9  \\
daubechies           &     18.8 &       NA &       NA &       NA  \\
haralick             &    233.1 &       NA &   7760.7 &       NA  \\
\bottomrule
\end{tabular}
\caption{Efficiency Results for mahotas, pymorph, scikits-image, and openCV
(through Python wrappers). Shown are values as multiples of the time that
\code{numpy.max(image)} takes to compute the maximum pixel value in the image
(all operations are over the same image). For scikits-image, features on the
grey-scale cooccurrence matrix were used instead of Haralick features, which it
does not support. In the case of \emph{median filter}, the radius of the
structuring element is shown in parentheses. NA stands for ``Not Available.''}
\label{tab:efficiency} \end{table}

Table~\ref{tab:efficiency} shows timings for different operations. These were
normalized to multiples of the time it takes to go over the image and find its
maximum pixel value (this was done using \code{numpy.max(image)}). The
measurements shown were obtained on an Intel 64~bit system, running Ubuntu
Linux. However, due to the normalization, measurements obtained on another
system (Intel 32~bits running Mac OS) were qualitatively similar.

The comparison is against Pymorph~\citep{DougLotu:03}, which is a pure Python
implementation of some of the same functions; scikits-image, which is a similar
project to mahotas, but with a heavier emphasis on the use of
Cython~\citep{Cython}; and OpenCV, which is a \cpp{} library with automatically
generated Python wrappers.

OpenCV is the fastest library, but this comes at the cost of some flexibility.
Arguments to its functions must be of the exact expected type and it is
possible to crash the interpreter if types do match the expected type (in the
other libraries, including mahotas, all types are checked and an exception is
generated which can be caught by user code). This is particularly relevant for
interactive use as the user is often exploring and is willing to pay the speed
cost of a few extra type checks and conversions to avoid a hard-crash.

Among the other libraries, mahotas is the fastest. Pymorph, even though it is
implemented in Python only, intelligently uses arithmetic operations for
morphological operation and can be very fast. However, for more complex
methods, such as watershed; its pure Python approach is very inefficient. The
one exception is that median filtering with a large structuring element is
faster in scikit-image. In fact, that library uses an algorithm with better
asymptotic behavior for this operation.

\subsection{Distribution and Installation}

In keeping with the philosophy of blending in with the ecosystem, Mahotas uses
the standard Python build machinery and distribution channels. Building and
installing from source code is done using
\begin{verbatim}
python setup.py install
\end{verbatim}
Alternatively, Python based package managers (such as \texttt{easy\_install} or
\texttt{pip}) can be used (mahotas works well with these systems).

\subsection{Quality Control}

Mahotas includes a complete automated suite of unit tests, which tests all
functionality and include several regression tests. There are no known bugs in
version~1.0. In fact, no releases have ever been performed with known bugs.
Naturally, bugs were, occasionally, discovered in released versions, but
corrected before the next release.

The development is completely open-source and development versions are
available. Many users have submitted bug reports and fixes.

\section{Availability}

\textbf{Operating system}\\
Mahotas runs and is used on different versions of Unix (including Linux, SunOS,
and FreeBSD), Mac OS X, and Windows.\footnote{Christoph Gohlke has been
instrumental in providing Windows packages as well as several fixes for that
platform.}

\textbf{Programming language}\\
Mahotas works in Python (minimal version is~2.5, but mahotas works with all
more recent versions, including version in the Python~3 series).

\textbf{Additional system requirements}\\
None at runtime. Compilation from source requires a \cpp{} compiler.

\textbf{Dependencies}\\
It requires numpy to be present and installed.

\textbf{List of contributors}\\
Luis Pedro Coelho (Carnegie Mellon University and Instituto de Medicina
Molecular), Zachary Pincus (Stanford University), Peter J. Verveer (European
Molecular Biology Laboratory), Davis King (Northrop Grumman ES), Robert Webb
(Carnegie Mellon University), Matthew Goodman (University of Texas at Austin),
K.-Michael Aye (University of Bern), Rita Sim\~{o}es (University of Twente),
Joe Kington (University of Wisconsin), Christoph Gohlke (University of
California, Irvine), Lukas Bossard (ETH Zurich), and Sandro Knauss (University
of Bremen).

\subsubsection{Software location}

\textbf{Code repository}\\
\textit{Name}: Github\\
\textit{Identifier}: https://github.com/luispedro/mahotas\\
\textit{Licence}: MIT\\
\textit{Date published}: Since 2010 as mahotas. Some of the code had been
previously made available under other names.

\section{Reuse Potential}

Originally, this code was developed in the context of cellular image analysis.
However, none of the functionality is specific to this context and many
computer vision pipelines can make use of it.

This package (and earlier versions of it) have been used by
myself~\citep{Coelho2009,Coelho2010a} and close collaborators in several
publications~\citep{omerosearcher}. Other groups have used in published work,
both in cell image analysis~\citep{CYTO:CYTO22034} and in other
areas~\citep{springerlink:10.1007/978-3-642-32335-5_2}.

\section{Discussion}

Python is an excellent language for scientific programming because of the
inherent properties of the language and because of the infrastructure that has
been built around the numpy project. Mahotas works in this environment to
provide the user with image analysis and computer vision functionality.

Mahotas does not include machine learning related functionality, such as
$k$-means clustering or classification methods. This is the result of an
explicit design decision. Specialised machine learning packages for Python
already
exist~\citep{Pedregosa:2011:SML:2078183.2078195,springerlink:10.1007/978-3-540-30116-5_58,Schaul:2010:PYB:1756006.1756030,Sonnenburg:2010:SML:1756006.1859911}.
A good classification system can benefit both computer vision users
and others. As these projects all use Numpy arrays as their data types, it is
easy to use functionality from the different project seamlessly (no copying of
data is necessary).

Mahotas is implemented in \cpp{}, as the standard Python interpreter is too
slow for a direct Python implementation. However, all of the Python interface
code is hand-written, as opposed to using automatic interface generators like
Swig~\citep{Beazley2003599}. This is more work initially, but the end result is
of much higher quality, especially when it comes to giving useful error
messages (e.g., when a type mismatch occurs, an automatic system will often be
forced to resort to a generic message as it does not have any knowledge of what
the arguments mean besides their automatically inferred types).

Mahotas has been available in the Python Package Index since April~2010 and has
been downloaded over 40,000~times. This does not include any downloads from
other sources. Mahotas includes a full test suite. There are no known bugs.

\subsection*{Acknowledgements}

Mahotas includes code ported and incorporated from other projects. Initially,
it was used in reproducing the functionality in the Subcellular Location Image
Classifier (SLIC) tool from Robert F.\ Murphy's
Lab~\citep{springerlink:10.1007/978-0-387-45524-2_47} and the initial versions
of mahotas were designed explicitly to support that functionality. The
\textsc{surf} implementation is a port from the code from
\textit{dlib},\footnote{Dlib's webpage is at \url{http://dlib.net}.} a very
good \cpp{} library by Davis King. I also gleaned some insight into the
implementation of these features from Christopher Evan's OpenSURF library and
its documentation \citep{evans2009}.\footnote{OpenSURF is available at
\url{http://www.chrisevansdev.com/computer-vision-opensurf.html}, where several
documents describe details of the implementation.} The code which interfaces
with the FreeImage library, was written by Zachary Pincus and some of the
support code was written by Peter J. Verveer for the \code{scipy.ndimage}
project. All of these contributions were integrated while respecting the
software licenses under which the original code had been released. Robert Webb,
a summer student at Carnegie Mellon University, worked with me on the initial
local binary patterns implementation. Finally, I thank the several users who
have reported bugs, submitted fixes, and participated on the project
mailing list.

St\'efan van der Walt and Andreas M\"uller offered helpful comments on a draft
version of this manuscript.

\textbf{Funding}: I was supported in my work by the Funda\c c\~{a}o para a
Ci\^encia e Tecnologia (grants SFRH/BD/37535/2007 and
PTDC/SAU-GMG/115652/2008), by NIH grant GM078622 (to Robert F.\ Murphy), by a
grant from the Scaife Foundation, by the HHMI Interfaces Initiative, and by a
grant from the Siebel Scholars Foundation.

\bibliography{references}
\end{document}